\documentclass[11pt,a4paper]{article}
\usepackage[hyperref]{emnlp2020}
\usepackage{times}
\usepackage{latexsym}
\usepackage{xcolor}
\usepackage{amsmath}
\usepackage{amsfonts}
\usepackage{graphicx}
\usepackage{algorithm}  
\usepackage{algorithmicx}  
\usepackage{algpseudocode}

\usepackage{booktabs}
\usepackage{multirow}
\usepackage{paralist}

\usepackage{microtype}

\aclfinalcopy 

\title{DGST: a Dual-Generator Network for Text Style Transfer}

\author{Xiao Li\textsuperscript{$\spadesuit$}, 
Guanyi Chen\textsuperscript{$\heartsuit$}, Chenghua Lin\textsuperscript{$\spadesuit$}\thanks{~~Corresponding author}~, Ruizhe Li\textsuperscript{$\spadesuit$}\\
\textsuperscript{$\spadesuit$}Department of Computer Science, University of Sheffield\\
\textsuperscript{$\heartsuit$}Department of Information and Computing Sciences, Utrecht University\\
\texttt{\{xiao.li, c.lin, r.li\}@sheffield.ac.uk, 
g.chen@uu.nl}}
\date{}

\begin{document}
\maketitle
\begin{abstract}

We propose DGST, a novel and simple Dual-Generator network architecture for text Style Transfer. 
Our model employs two generators only, and  does not rely on any discriminators or parallel corpus for training.
Both quantitative and qualitative experiments on the Yelp and IMDb datasets show that  our model gives competitive performance compared to several strong baselines with more complicated architecture designs.

\end{abstract}

\section{Introduction}

Attribute style transfer is a task which seeks to change a stylistic attribute of text, while preserving its attribute-independent information. Sentiment transfer is a typical example of such kind, which focuses on controlling the sentiment polarity of the input text~\cite{shen2017style}. Given a review \emph{``the service was very poor''}, a successful sentiment transferrer should covert the negative sentiment of the input to positive (e.g., replacing the phrase \emph{``very poor''} with \emph{``pretty good''}), while keeping all other information unchanged (e.g., the aspect \emph{``service''} should not being changed to \emph{``food''}).

Without supervised signals from parallel data, a transferrer must be supervised in a way to ensure that the generated texts belongs to a certain style category (i.e., transfer intensity).
There is a growing body of studies to intensify the target style by means of adversarial training~\citep{fu2018style}, variational autoencoder~\cite{john2019disentangled,li2019stable,fang2019implicit}, generative adversarial nets~\citep{shen2017style, zhao2017adversarially, yang2018unsupervised}, or subspace matrix projection~\citep{msp}

Furthermore, in order to boost the preservation of non-attribute information during style transformation, some works explicitly focus on modifying sentiment words, which is so-called the ``pivot word''~\citep{li2018delete, wu2019mask}. 
There are also works which add extra components for constraining the content from being changed too much. These include models like autoencoder~\citep{lample2018multipleattribute, dai2019style}, part-of-speech preservation, and the content conditional language model~\citep{tian2018structured}.
In order to achieve high-quality style transfer, existing works normally resort to adding additional inner or outer structures such as additional adversarial networks or data pre-processing steps (e.g. generating pseudo-parallel corpora). 
This inevitably increases the complexity of the model and raises the bar of training data requirement.

In this paper, we propose a novel and simple model architecture for text style transfer, which employs two generators only. In contrast to some of the dominant approaches to style transfer such as CycleGAN~\citep{zhu2017unpaired}, our model does not employ any discriminators and yet can be trained without requiring any parallel corpus.
We achieve this by developing a novel sentence noisification approach called \textit{neighbourhood sampling}, which can introduce noise to each input sentence dynamically. The nosified sentences are then used to train our style transferrers in the way similar to the training of denoising autoencoders~\citep{vincent2008extracting}. Both quantitative and qualitative evaluation on the Yelp and IMDb benchmark datasets show that DGST gives competitive performance compared to several strong baselines which have more complicated model design. 
The code of DGST is available at: \url{https://xiao.ac/proj/dgst}.

\section{Methodology}\label{sec:method}

Suppose we have two non-parallel corpora 
 $X$ and $Y$ with style $S_x$ and $S_y$, the goal is training two transferrers, each of which can (i) transfer a sentence from one style (either $S_x$ and $S_y$) to another (i.e., transfer intensity); and (ii) preserve the style-independent context during the transformation (i.e., preservation). Specifically, we denote the two transferrers $f$ and $g$. $f:\mathcal{X} \to \mathcal{Y}$ transfers a sentence $x \in \mathcal{X}$ with style $S_x$ to $y^*$ with style $S_y$. Likewise,  $g: \mathcal{Y} \to \mathcal{X}$ transfers a sentence $y \in \mathcal{Y}$ with style $S_y$ to $x^*$ with $S_x$.
To obtain good style transfer performance, $f$ and $g$ need to achieve both a high transfer intensity and a high preservation, which can be formulated as follows:
\begin{align}
& \forall x, \forall x' \in \mathcal{X}, ~ \forall y, \forall y' \in \mathcal{Y} \notag \\
& y^*=f(x) \in \mathcal{Y}, x^*=g(y) \in \mathcal{X} \label{eq:1a}\\
& D(y^*||x) \leq D(y'||x), D(x^*||y) \leq D(x'||y) \label{eq:1b}
\end{align}

Here $D(x||y)$ is a function that measures the abstract distance between sentences in terms of the minimum edit distance, where the editing operations $\Phi$ includes word-level replacement, insertion, and deletion (i.e., the Hamming distance or the Levenshtein distance). On the one hand, Eq.~\ref{eq:1a} requires the transferred text should fall within the target style spaces (i.e., $\mathcal{X}$ or $\mathcal{Y}$). On the other hand, Eq.~\ref{eq:1b} constrains the transferred text from changing too much, i.e., to preserve the style-independent information.

\begin{figure}[bt]
\centering
\includegraphics[width=2.8in]{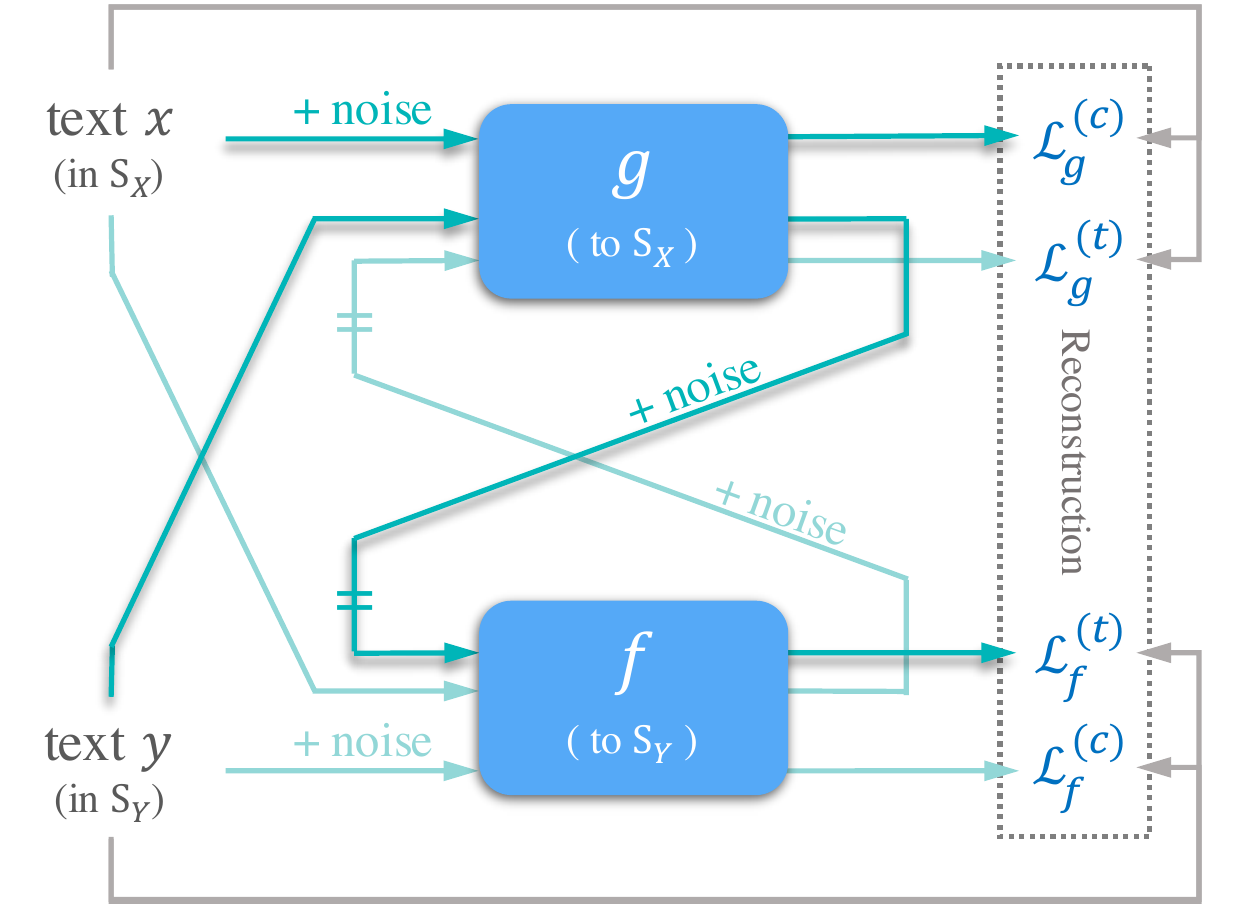}
\caption{The general architecture of DGST, in which ``='' means no  back-propagation of gradients.}
\label{fig:model}
\end{figure}

Inspired by CycleGAN~\citep{zhu2017unpaired}, our model (sketched in Figure~\ref{fig:model}) is trained by a cyclic process: for each transferrer, a text is transferred to the target style, and then back-transferred to the source style using another transferrer.
In order to transfer a sentence to a target style while preserving the style-independent information, we formulate two sets of training objectives: one set ensures that the generated sentences is preserved as much as possible (detailed in \S \ref{sec:preserve}) and the other set is responsible for transferring the input text to the target style (detailed in \S \ref{sec:transfer}).

\subsection{Preserving the Content of Input Text} \label{sec:preserve}%

This section discusses our loss function which enforces our transferrers to preserve the style-independent information of the input.  
A common solution to this problem is to use the reconstruction loss of the autoencoders~\citep{dai2019style}, which is also known as the identity loss~\citep{zhu2017unpaired}. However, too much emphasis on preserving the content would hinder the style transferring ability of the transferrers. To balance our model's capability in content preservation and transfer intensity, we instead first train our transferrers in the way of training denoising autoencoders~\citep[DAE,][]{vincent2008extracting}, which has been proved to help preserving the style independent content of  input text~\citep{shen2019educating}.
More specifically, we train $f$ (or $g$; we use $f$ as an example in the rest of this section) by feeding it with a noisy sentence $\mathring{y}$ as input, where $\mathring{y}$ is noisified from $y \in \mathcal{Y}$ and $f$ is expected to reconstruct $y$.

Different from previous works which use DAE in style transfer or MT~\citep{artetxe2017unsupervised,lample2018multipleattribute}, 
we propose a novel sentence noisification approach, named \textit{neighbourhood sampling}, 
which introduces noise to each sentence dynamically.
For a sentence $y$, we define $U_\alpha(y, \gamma)$ as a neighbourhood of $y$, which is a set of sentences consisting of $y$ and all variations of noisified $y$ with the same noise intensity $\gamma$ (which will be explained later).
The size of the neighbourhood $U_\alpha(y, \gamma)$ is determined by the proportion (denoted by $m$) of tokens in $y$ that are modified using the editing operations in $\Phi$.
Here the proportion $m$ is sampled from a Folded Normal Distribution $\mathcal{F}$. We hereby define that the average value of $m$ (i.e., the mean of $\mathcal{F}$) is the noise intensity $\gamma$. Formally,  $m$ is defined as:
\begin{equation}\label{eq:foldn}
m \sim \mathcal{F}(m';\gamma) =\frac{2}{\pi \gamma}e^{-\frac{m'^2}{\pi \gamma^2}}
\end{equation}
That said, a neighbourhood $U_\alpha(y, \gamma)$ would be constructed using $y$ and all sentences that are created by modifying $(m \times \text{length}(y))$ words in $y$, from which we sample $\mathring{y}$, i.e., a noisified sentence of $y$: $\mathring{y} \sim U_\alpha(y, \gamma)$. Analogously, we could also construct a neighbourhood $U_\beta(x, \gamma)$ for $x \in \mathcal{X}$ and sample $\mathring{x}$ from it. Using these noisified data as inputs, we then train our transferrers $f$ and $g$ in the way of DAE by optimising the following reconstruction objectives:
\begin{equation}\label{eq:correct}
\begin{aligned}
\mathcal{L}_f^{(c)} & = \mathbb{E}_{y\sim Y,\mathring{y}\sim U_\alpha (y,\gamma)} D(y||f(\mathring{y}))   \\
\mathcal{L}_g^{(c)} & = \mathbb{E}_{x\sim X,\mathring{x}\sim U_\beta (x,\gamma)} D(x||g(\mathring{x}))  
\end{aligned}
\end{equation}
With Eq.~\ref{eq:correct}, we essentially encourages the generator to preserve the input as much as possible.

\subsection{Transferring Text Styles} \label{sec:transfer}

Making use of non-parallel datasets, we train $f$ and $g$ in an iterative process. 
Let $M=\{g(y)|y \in Y\}$ be the range of  $g$ when the input is all sentences in the training set $Y$. Similarly, we can define  $N=\{f(x)|x \in X\}$. 
 During the training cycle of $f$, $g$ will be kept unchanged. 
We first feed each sentence $y$ ($y\in Y$) to $g$,  which tries to transfer $y$ to the target style $\mathcal X$ (i.e. ideally $x^*=g(y) \in \mathcal X$). In this way, we obtain $M$ which is composed of all $x^*$ for each $y\in Y$.
Next, we sample $\mathring{x}^*$ (a noised sentence of $x^*$) based on $x^*$ via the neighbourhood sampling, i.e., $\mathring{x}^* \sim U_\alpha(x^*,\gamma)=U_\alpha(g(y),\gamma)$. We use $\mathring M$ to represent the collection of $\mathring{x}^*$.
Similarly, we obtains $N$ and $\mathring N$ using the aforementioned procedures during the training cycle for $g$.

Instead of directly using the sentences from $X$ for training, we use $\mathring M$ to train $f$ by forcing $f$ to transfer each $\mathring x^*$ back to the corresponding original $y$. In parallel, $\mathring N$ is utilised to train $g$. 
We represent the aforementioned operation as the \textit{transfer objective}. 
\begin{equation}\label{eq:transfer}
\begin{aligned}
\mathcal{L}_f^{(t)} & = \mathbb{E}_{\alpha,y\sim Y,\mathring{x}^*\sim U_\alpha (g(y),\gamma)} D(y||f(\mathring{x}^*)) \\
\mathcal{L}_g^{(t)} & =  \mathbb{E}_{\beta,x\sim X,\mathring{y}^*\sim U_\beta (f(x),\gamma)} D(x||g(\mathring{y}^*))
\end{aligned}
\end{equation}
The main difference between Eq.~\ref{eq:correct} and Eq.~\ref{eq:transfer} is how $U_\alpha (\cdot,\gamma)$ and $U_\beta (\cdot,\gamma)$ are constructed, i.e., $U_\alpha (y,\gamma)$ and $U_\beta (x,\gamma)$ in Eq.~\ref{eq:correct} compared to $U_\alpha (g(y),\gamma)$ and $U_\beta (f(x),\gamma)$ in Eq.~\ref{eq:transfer}.
Finally, the overall loss of DGST is the sum of the four partial losses:
\begin{equation}
\mathcal{L} = 
\mathcal{L}_f^{(c)}
+\mathcal{L}_f^{(t)}
+\mathcal{L}_g^{(c)}
+\mathcal{L}_g^{(t)}
\end{equation}

During optimisation, we freeze $g$ when optimising $f$, and vice versa.
Also with the reconstruction objective, $x^*$ must to be sampled first, and then passed $\mathring{x}^\ast$ into $f$; in contrast, it is not necessary to sample according to $y$ when we obtain $x^\ast = g(y)$.

\section{Experiment}

\begin{table}[thb!]
\small
\centering
 \begin{tabular}{lccccc}
  \toprule
  \multirow{2}{*}{Dataset} & \multicolumn{2}{c}{Yelp} & & \multicolumn{2}{c}{IMDb} \\ \cline{2-3} \cline{5-6}
   & Positive & Negative & & Positive & Negative\\
  \midrule
Train & 266,041 & 177,218 & & 178,869 & 187,597 \\
Dev & 2,000 & 2,000 & & 2,000 & 2,000 \\
Test & 500 & 500 & & 1,000 & 1,000 \\
  \bottomrule
 \end{tabular}
 \caption{Statistics of Datasets.}
\label{tab:state}
\end{table}

\begin{table*}[tb]
\centering \small
 \begin{tabular}{lcccccc}
  \toprule
  \multirow{2}{*}{Model} & \multicolumn{3}{c}{Yelp} & & \multicolumn{2}{c}{IMDb} \\ \cline{2-4} \cline{6-7}
   & acc. & \emph{ref}-BLEU & \emph{self}-BLEU & & acc. & \emph{self}-BLEU\\
  \midrule
  RetrieveOnly~\citep{li2018delete} & 92.6 & 0.4 & 0.7 & & n/a & n/a \\
  TemplateBased~\citep{li2018delete} & 84.3 & 13.7 & 44.1 & & n/a & n/a \\
  DeleteOnly~\citep{li2018delete} & 85.7 & 9.7 & 28.6 & & n/a & n/a \\
  DeleteAndRetrieve~\citep{li2018delete} & 87.7 & 10.4 & 29.1 & & 55.8 & 55.4 \\
  ControlledGen~\citep{hu2017toward} & 88.8 & 14.3 & 45.7 & & 94.1 & 62.1 \\
  CycleRL~\citep{xu2018unpaired} & 88.0 & 2.8 & 7.2 & & \textbf{97.8} & 4.9 \\
  StyleTransformer (Conditional)~\citep{dai2019style} & \textbf{93.7} & 17.1 & 45.3 & & 86.6 & 66.2 \\
  StyleTransformer (Multi-Class)~\citep{dai2019style} & 87.7 & \textbf{20.3} & \textbf{54.9} & & 80.3 & \textbf{70.5} \\
  \midrule
  DGST & 88.0 & 18.7 & 54.5 &  & 70.1 & 70.2 \\
  \bottomrule
 \end{tabular}
 \caption{Automatic evaluation results on Yelp and IMDb corpora, most of which are from \citet{dai2019style}.}
\label{tab:auto}
\end{table*}

\begin{table*}[tb]
\small
\centering
 \begin{tabular}{r|l}
  \toprule
  \textbf{Yelp} & \textbf{positive $\rightarrow$ negative} \\
  \midrule
  input & this golf club is one of the best in my opinion . \\
  output & this golf club is one of the worst in my opinion . \\ 
  \midrule
  input & i definitely recommend this place to others ! \\
  output & i do not recommend this to anyone ! \\
  \toprule
  \textbf{Yelp} & \textbf{negative $\rightarrow$ positive} \\
  \midrule
  input & the garlic bread was bland and cold . \\
  output & the garlic bread was tasty and fresh . \\
  \midrule
  input & my dish was pretty salty and could barely taste the garlic crab . \\
  output & my dish was pretty good and could even taste the garlic crab . \\
  \midrule
  \textbf{IMDb} & \textbf{positive $\rightarrow$ negative} \\
  \midrule
  input & a timeless classic , one of the best films of all time . \\
  output & a complete disaster , one of the worst films of all time . \\
  \midrule
  input & and movie is totally backed up by the excellent music both in background and in songs by monty . \\
  output & the movie is totally messed up by the awful music both in background and in songs by chimps . \\
  \toprule
  \textbf{IMDb} & \textbf{negative $\rightarrow$ positive} \\
  \midrule
  input & this one is definitely one for my `` worst movies ever '' list . \\
  output & this one is definitely one of my `` best movies ever '' list . \\
  \midrule
  input & i found this movie puerile and silly , as well as predictable . \\
  output & i found this movie credible and funny , as well as tragic . \\
  \bottomrule
 \end{tabular}
 \caption{Example results from our model for the sentiment style transfer on the Yelp and IMDb datasets.}
\label{tab:transferexample}
\end{table*}

\begin{table*}[tb]
\resizebox{\linewidth}{!}{
\small
\centering
 \begin{tabular}{r|l|l}
  \toprule
  & \textbf{positive $\rightarrow$ negative} & \textbf{negative $\rightarrow$ positive} \\
  \midrule
  \textit{input} & \textit{it is a cool place , with lots to see and try .} & \textit{so , that was my one and only time ordering the benedict there .} \\
  \textbf{full-model} & it is a sad place , with lots to see and something . & so , that was my one and best time in the shopping there . \\
  no-rec & no no , , \_num\_ . & so , that was my one and time time over the there there . \\
  rec-no-noise & it is a cool place , with me to see and try . & service was very friendly . \\
  no-tran & it is a loud place , with lots to try and see . & so , that was my only and first visit ordering the there ) . \\
  tran-no-noise & it is a noisy place , with lots to try and see . & so , that was my one and time time ordering the ordering there . \\
  pre-noise & it is a cool place , with lots to see to try . & so , that 's one one and my only the the day there . \\
  \midrule
  \textit{input} & \textit{it is the most authentic thai in the valley .} & \textit{even if i was insanely drunk , i could n't force this pizza down .} \\
  \textbf{full-model} & it is the most overrated thai in the valley . & even if i was n't hungry , i 'll definitely enjoy this pizza here . \\
  no-rec & i was in the the the the food . & she was perfect . \\
  rec-no-noise & it is the most authentic thai in the valley . & even if i was n't , , i could n't recommend this pizza . . \\
  no-tran & it is the most authentic thai in the valley . & even if i was n't , , i could n't get this pizza down . \\ 
  tran-no-noise & it is the most common thai in the valley . & even if i was hungry hungry , i could n't love this pizza shop . \\
  pre-noise & it is the most thai thai in the valley . & even if i was n't hungry , i could n't recommend this pizza down . \\
  \bottomrule
 \end{tabular}
 }
 \caption{Example transferred from the ablation study.}
\label{tab:soexample}
\end{table*}

\subsection{Setup}
\noindent \textbf{Dataset.}~~We evaluated our model 
on two benchmark datasets, namely, the Yelp review dataset (Yelp), which consists of restaurants and business reviews together with their sentiment polarity (i.e., positive or negative), and  the IMDb Movie Review Dataset (IMDb), which consists of online movie reviews. For Yelp, we split the dataset following \citet{li2018delete}, who also provided human produced reference sentences for evaluation. For IMDb, we follow the pre-processing and data splitting protocol of \citet{dai2019style}. Detailed dataset statistics is given in  Table~\ref{tab:state}.

\noindent \textbf{Evaluation Protocol.}~~Following the standard evaluation practice, we evaluate the performance of our model on the textual style transfer task from two aspects: 
(1) \textbf{Transfer Intensity}: a style classifier is employed for quantifying the intensity of the transferred text. In our work, we use FastText \citep{joulin2017bag} trained on the training set of Yelp;
(2) \textbf{Content Preservation}: to validate whether the style-independent context is preserved by the transferrer, we calculate \emph{self}-BLEU, which computes a BLEU score~\citep{papineni2002bleu} by comparing inputs and outputs of a system. A higher \emph{self}-BLEU score indicates more tokens from the sources are retained, henceforth, better preservation of the contents. In addition, we also use \emph{ref}-BLEU, which compares the system outputs and the references written by human beings.

\subsection{Experimental Results}
In our experiment, the two transferrers ($f$ and $g$) are  Stacked BiLSTM-based sequence-to-sequence models, i.e., both 4-layer BiLSTM for the encoder and decoder. The noise intensity $\gamma$ is set to 0.3 in the first 50 epochs and 0.03 in the following epochs.

As shown in Table~\ref{tab:auto}, for the Yelp dataset our model defeats  all  baselines models (apart from StyleTransformer (Multi-Class)) on both \emph{ref}-BLEU and \emph{self}-BLEU. 
In addition, as shown in Table~\ref{tab:auto}, our model works remarkably well on both transfer intensity and preservation without requiring adversarial training or reinforcement learning, or external offline sentiment classifiers (as in \citet{dai2019style}). Besides, the current version of our model is built upon fundamental BiLSTM, which is a likely explanation of why we lose to the SOTA (i.e., StyleTransformer (Multi-Class)) for a small margin, which are based on the Transformer architecture~\citep{vaswani2017attention} with much higher capacity. For the IMDb dataset, comparing to other systems, our model obtained moderate accuracy but competitive \emph{self-}BLEU score (70.2), i.e., slightly lower than StyleTransformer. 
Table~\ref{tab:transferexample} lists several examples for style transfer in sentiment for both datasets. By examining the results, we can see that DGST is quite effective in transferring the sentiment polarity of the input sentence while maintaining the non-sentiment information.

\begin{table}[t]
\small
\centering
 \begin{tabular}{l|cc}
  \toprule
  Model Variants & \emph{self}-BLEU & acc. \\
  \midrule
  no-rec    & 0.0 & 98.9 \\
  rec-no-noise & 41.9 & 73.1 \\
  no-tran    & 98.0 & 4.2 \\
  tran-no-noise & 35.6 & 82.9 \\
  pre-noise      & 38.9 & 76.8 \\
  \midrule
  full-model   & 37.2 & 86.3 \\
  \bottomrule
 \end{tabular}
 \caption{Evaluation results for the ablation study.}
\label{tab:so}
\end{table}

\subsection{Ablation Study}
To confirm the validity of our model, we did an ablation study on Yelp by eliminating or modifying a certain component (e.g., objective functions, or sampling neighbourhood). We tested the following variations: 1) \textbf{full-model}: the proposed model; 2) \textbf{no-tran}: the model without the transfer objective; 3) \textbf{no-rec}: the model without the reconstruction objective; 4) \textbf{rec-no-noise}: the model adding no noise when optimising the reconstruction objective; 5) \textbf{tran-no-noise}: the model adding no noise when optimising the transfer objective; 6) \textbf{pre-noise}: the model trained by adding noise to $y$ first and then feeding the nosified sentences $\mathring{y}$ to $g$ (or $\mathring{x}$ to $f$) in Eq.~\ref{eq:transfer}. 
In this study, the transferrers are the simplest LSTM-based sequence-to-sequence models. The hidden size and $\gamma$ are set to 256 and 0.3, respectively.

\noindent \textbf{Results.} Table~\ref{tab:so} depicts the results of the ablation study. As we can see, eliminating the reconstruction or transfer objectives would damage preservation and transfer intensity, respectively. As for the use of noise, the results of the rec-no-noise model shows that the noise in the reconstruction objective helps balance our model’s ability in content preservation and transfer intensity. For the transfer objective, omitting noise (tran-no-sp) would reduce the transfer intensity while placing noise in the wrong position (pre-noise) reduces it yet again.

\noindent \textbf{Case Study.} Transferred sentences produced by each model variant in the ablation study are listed in Table~\ref{tab:soexample}. The model without correction objective (no-corr)  collapsed and as a result it generates irrelevant sentences to the inputs most of the time. When neighbourhood sampling is dropped in either corrective or transfer objectives, the transfer intensity is reduced. 
These models, including rec-no-noise, tran-no-noise, and pre-noise, tend to substitute random words, and result in reduced transfer intensity (i.e., style  words  are  either not  modified or still express the same sentiment after modification) and preservation. For example, when transferring from negative to positive, rec-no-noise replace \emph{``force''} to \emph{``recommend''} resulting \emph{``I couldn't recommend this pizza''}, which is still a negative review.

\section{Conclusion}
In this paper, we propose a novel and simple dual-generator network architecture for text style transfer, which does not rely on any discriminators or  parallel corpus for training. 
Extensive experiments on two public datasets show that  our model yields competitive performance compared to several strong baselines, despite of our simpler model architecture design.

\section*{Acknowledgements}
We would like to thank all the anonymous reviewers for their insightful comments. This work is supported by the award made by the
UK Engineering and Physical Sciences Research
Council (Grant number: EP/P011829/1).

\bibliography{emnlp2020}

\begin{thebibliography}{22}
\expandafter\ifx\csname natexlab\endcsname\relax\def\natexlab#1{#1}\fi

\bibitem[{Artetxe et~al.(2018)Artetxe, Labaka, Agirre, and
  Cho}]{artetxe2017unsupervised}
Mikel Artetxe, Gorka Labaka, Eneko Agirre, and Kyunghyun Cho. 2018.
\newblock \href {https://openreview.net/forum?id=Sy2ogebAW} {Unsupervised
  neural machine translation}.
\newblock In \emph{International Conference on Learning Representations}.

\bibitem[{Dai et~al.(2019)Dai, Liang, Qiu, and Huang}]{dai2019style}
Ning Dai, Jianze Liang, Xipeng Qiu, and Xuanjing Huang. 2019.
\newblock \href {https://doi.org/10.18653/v1/P19-1601} {Style transformer:
  Unpaired text style transfer without disentangled latent representation}.
\newblock In \emph{Proceedings of the 57th Annual Meeting of the Association
  for Computational Linguistics}, pages 5997--6007, Florence, Italy.
  Association for Computational Linguistics.

\bibitem[{Fang et~al.(2019)Fang, Li, Gao, Dong, and Chen}]{fang2019implicit}
Le~Fang, Chunyuan Li, Jianfeng Gao, Wen Dong, and Changyou Chen. 2019.
\newblock Implicit deep latent variable models for text generation.
\newblock In \emph{Proceedings of the 2019 Conference on Empirical Methods in
  Natural Language Processing and the 9th International Joint Conference on
  Natural Language Processing (EMNLP-IJCNLP)}, pages 3937--3947.

\bibitem[{Fu et~al.(2018)Fu, Tan, Peng, Zhao, and Yan}]{fu2018style}
Zhenxin Fu, Xiaoye Tan, Nanyun Peng, Dongyan Zhao, and Rui Yan. 2018.
\newblock Style transfer in text: Exploration and evaluation.
\newblock In \emph{Thirty-Second AAAI Conference on Artificial Intelligence}.

\bibitem[{Hu et~al.(2017)Hu, Yang, Liang, Salakhutdinov, and
  Xing}]{hu2017toward}
Zhiting Hu, Zichao Yang, Xiaodan Liang, Ruslan Salakhutdinov, and Eric~P Xing.
  2017.
\newblock Toward controlled generation of text.
\newblock In \emph{Proceedings of the 34th International Conference on Machine
  Learning-Volume 70}, pages 1587--1596. JMLR. org.

\bibitem[{John et~al.(2019)John, Mou, Bahuleyan, and
  Vechtomova}]{john2019disentangled}
Vineet John, Lili Mou, Hareesh Bahuleyan, and Olga Vechtomova. 2019.
\newblock Disentangled representation learning for non-parallel text style
  transfer.
\newblock In \emph{Proceedings of the 57th Annual Meeting of the Association
  for Computational Linguistics}, pages 424--434.

\bibitem[{Joulin et~al.(2017)Joulin, Grave, Bojanowski, and
  Mikolov}]{joulin2017bag}
Armand Joulin, Edouard Grave, Piotr Bojanowski, and Tomas Mikolov. 2017.
\newblock Bag of tricks for efficient text classification.
\newblock In \emph{Proceedings of the 15th Conference of the European Chapter
  of the Association for Computational Linguistics: Volume 2, Short Papers},
  pages 427--431. Association for Computational Linguistics.

\bibitem[{Lample et~al.(2019)Lample, Subramanian, Smith, Denoyer, Ranzato, and
  Boureau}]{lample2018multipleattribute}
Guillaume Lample, Sandeep Subramanian, Eric Smith, Ludovic Denoyer,
  Marc'Aurelio Ranzato, and Y-Lan Boureau. 2019.
\newblock Multiple-attribute text rewriting.
\newblock In \emph{International Conference on Learning Representations}.

\bibitem[{Li et~al.(2018)Li, Jia, He, and Liang}]{li2018delete}
Juncen Li, Robin Jia, He~He, and Percy Liang. 2018.
\newblock \href {https://doi.org/10.18653/v1/N18-1169} {Delete, retrieve,
  generate: a simple approach to sentiment and style transfer}.
\newblock In \emph{Proceedings of the 2018 Conference of the North {A}merican
  Chapter of the Association for Computational Linguistics: Human Language
  Technologies, Volume 1 (Long Papers)}, pages 1865--1874, New Orleans,
  Louisiana. Association for Computational Linguistics.

\bibitem[{Li et~al.(2019)Li, Li, Lin, Collinson, and Mao}]{li2019stable}
Ruizhe Li, Xiao Li, Chenghua Lin, Matthew Collinson, and Rui Mao. 2019.
\newblock A stable variational autoencoder for text modelling.
\newblock In \emph{Proceedings of the 12th International Conference on Natural
  Language Generation}, pages 594--599.

\bibitem[{Li et~al.(2020)Li, Lin, Li, Wang, and Guerin}]{msp}
Xiao Li, Chenghua Lin, Ruizhe Li, Chaozheng Wang, and Frank Guerin. 2020.
\newblock Latent space factorisation and manipulation via matrix subspace
  projection.
\newblock In \emph{Proceedings of Machine Learning and Systems 2020}, pages
  3211--3221.

\bibitem[{Papineni et~al.(2002)Papineni, Roukos, Ward, and
  Zhu}]{papineni2002bleu}
Kishore Papineni, Salim Roukos, Todd Ward, and Wei-Jing Zhu. 2002.
\newblock Bleu: a method for automatic evaluation of machine translation.
\newblock In \emph{Proceedings of the 40th annual meeting on association for
  computational linguistics}, pages 311--318. Association for Computational
  Linguistics.

\bibitem[{Shen et~al.(2017)Shen, Lei, Barzilay, and Jaakkola}]{shen2017style}
Tianxiao Shen, Tao Lei, Regina Barzilay, and Tommi Jaakkola. 2017.
\newblock Style transfer from non-parallel text by cross-alignment.
\newblock In \emph{Advances in neural information processing systems}, pages
  6830--6841.

\bibitem[{Shen et~al.(2020)Shen, Mueller, Barzilay, and
  Jaakkola}]{shen2019educating}
Tianxiao Shen, Jonas Mueller, Regina Barzilay, and Tommi Jaakkola. 2020.
\newblock Educating text autoencoders: Latent representation guidance via
  denoising.
\newblock In \emph{Proceedings of Machine Learning and Systems 2020}, pages
  9129--9139.

\bibitem[{Tian et~al.(2018)Tian, Hu, and Yu}]{tian2018structured}
Youzhi Tian, Zhiting Hu, and Zhou Yu. 2018.
\newblock Structured content preservation for unsupervised text style transfer.
\newblock \emph{arXiv preprint arXiv:1810.06526}.

\bibitem[{Vaswani et~al.(2017)Vaswani, Shazeer, Parmar, Uszkoreit, Jones,
  Gomez, Kaiser, and Polosukhin}]{vaswani2017attention}
Ashish Vaswani, Noam Shazeer, Niki Parmar, Jakob Uszkoreit, Llion Jones,
  Aidan~N Gomez, {\L}ukasz Kaiser, and Illia Polosukhin. 2017.
\newblock Attention is all you need.
\newblock In \emph{Advances in neural information processing systems}, pages
  5998--6008.

\bibitem[{Vincent et~al.(2008)Vincent, Larochelle, Bengio, and
  Manzagol}]{vincent2008extracting}
Pascal Vincent, Hugo Larochelle, Yoshua Bengio, and Pierre-Antoine Manzagol.
  2008.
\newblock Extracting and composing robust features with denoising autoencoders.
\newblock In \emph{Proceedings of the 25th international conference on Machine
  learning}, pages 1096--1103.

\bibitem[{Wu et~al.(2019)Wu, Zhang, Zang, Han, and Hu}]{wu2019mask}
Xing Wu, Tao Zhang, Liangjun Zang, Jizhong Han, and Songlin Hu. 2019.
\newblock \href {https://doi.org/10.24963/ijcai.2019/732} {Mask and infill:
  Applying masked language model for sentiment transfer}.
\newblock In \emph{Proceedings of the Twenty-Eighth International Joint
  Conference on Artificial Intelligence, {IJCAI-19}}, pages 5271--5277.
  International Joint Conferences on Artificial Intelligence Organization.

\bibitem[{Xu et~al.(2018)Xu, Sun, Zeng, Zhang, Ren, Wang, and
  Li}]{xu2018unpaired}
Jingjing Xu, Xu~Sun, Qi~Zeng, Xiaodong Zhang, Xuancheng Ren, Houfeng Wang, and
  Wenjie Li. 2018.
\newblock \href {https://doi.org/10.18653/v1/P18-1090} {Unpaired
  sentiment-to-sentiment translation: A cycled reinforcement learning
  approach}.
\newblock In \emph{Proceedings of the 56th Annual Meeting of the Association
  for Computational Linguistics (Volume 1: Long Papers)}, pages 979--988,
  Melbourne, Australia. Association for Computational Linguistics.

\bibitem[{Yang et~al.(2018)Yang, Hu, Dyer, Xing, and
  Berg-Kirkpatrick}]{yang2018unsupervised}
Zichao Yang, Zhiting Hu, Chris Dyer, Eric~P Xing, and Taylor Berg-Kirkpatrick.
  2018.
\newblock Unsupervised text style transfer using language models as
  discriminators.
\newblock In \emph{Advances in Neural Information Processing Systems}, pages
  7287--7298.

\bibitem[{Zhao et~al.(2018)Zhao, Kim, Zhang, Rush, and
  LeCun}]{zhao2017adversarially}
Junbo Zhao, Yoon Kim, Kelly Zhang, Alexander Rush, and Yann LeCun. 2018.
\newblock \href {http://proceedings.mlr.press/v80/zhao18b.html} {Adversarially
  regularized autoencoders}.
\newblock volume~80 of \emph{Proceedings of Machine Learning Research}, pages
  5902--5911, Stockholmsmässan, Stockholm Sweden. PMLR.

\bibitem[{Zhu et~al.(2017)Zhu, Park, Isola, and Efros}]{zhu2017unpaired}
Jun-Yan Zhu, Taesung Park, Phillip Isola, and Alexei~A Efros. 2017.
\newblock Unpaired image-to-image translation using cycle-consistent
  adversarial networks.
\newblock In \emph{Proceedings of the IEEE international conference on computer
  vision}, pages 2223--2232.

\end{thebibliography}
\bibliographystyle{acl_natbib}

\end{document}